\pdfoutput=1

\documentclass[11pt]{article}

\usepackage[preprint]{coling}

\usepackage{times}
\usepackage{color}
\def\red#1{\textcolor[rgb]{1,0,0}{#1}}
\usepackage{multicol}
\usepackage{multirow}
\usepackage{graphicx}
\usepackage{amsfonts}
\usepackage{array}
\usepackage{booktabs}
\usepackage{bbding}
\usepackage{hyperref}
\usepackage{makecell}
\usepackage{amsmath}
\usepackage{latexsym}
\usepackage{float}
\usepackage{subfig}

\usepackage[T1]{fontenc}

\usepackage[utf8]{inputenc}

\usepackage{microtype}

\usepackage{inconsolata}

\usepackage{graphicx}

\title{EAMA : Entity-Aware Multimodal Alignment Based \\ Approach for News Image Captioning}

\author{Junzhe Zhang $\,$ Huixuan Zhang $\,$ Xunjian Yin $\,$ Xiaojun Wan \\
  Wangxuan Institute of Computer Technology, Peking University \\
  School of Electronics Engineering and Computer Science, Peking University\\
  \texttt{\{junzhezhang, zhanghuixuan\}@stu.pku.edu.cn} \\
  \texttt{\{xjyin, wanxiaojun\}@pku.edu.cn} \\}

\begin{document}
\maketitle
\begin{abstract}
News image captioning requires model to generate an informative caption rich in entities, with the news image and the associated news article. Current MLLMs still bear limitations in handling entity information in news image captioning tasks. Besides, generating high-quality news image captions requires a trade-off between sufficiency and conciseness of textual input information. To explore the potential of MLLMs and address problems we discovered, we propose \textit{EAMA}: an \textbf{E}ntity-\textbf{A}ware \textbf{M}ultimodal \textbf{A}lignment based approach for News Image Captioning. Our approach first aligns the MLLM with two extra alignment tasks: Entity-Aware Sentence Selection task and Entity Selection task, together with News Image Captioning task. The aligned MLLM will utilize the additional entity-related information extracted by itself to supplement the textual input while generating news image captions. Our approach achieves better results than all previous models on two mainstream news image captioning datasets.

\end{abstract}

\section{Introduction}

\begin{figure}[tbp]
    \centering
    \includegraphics[width=0.45\textwidth]{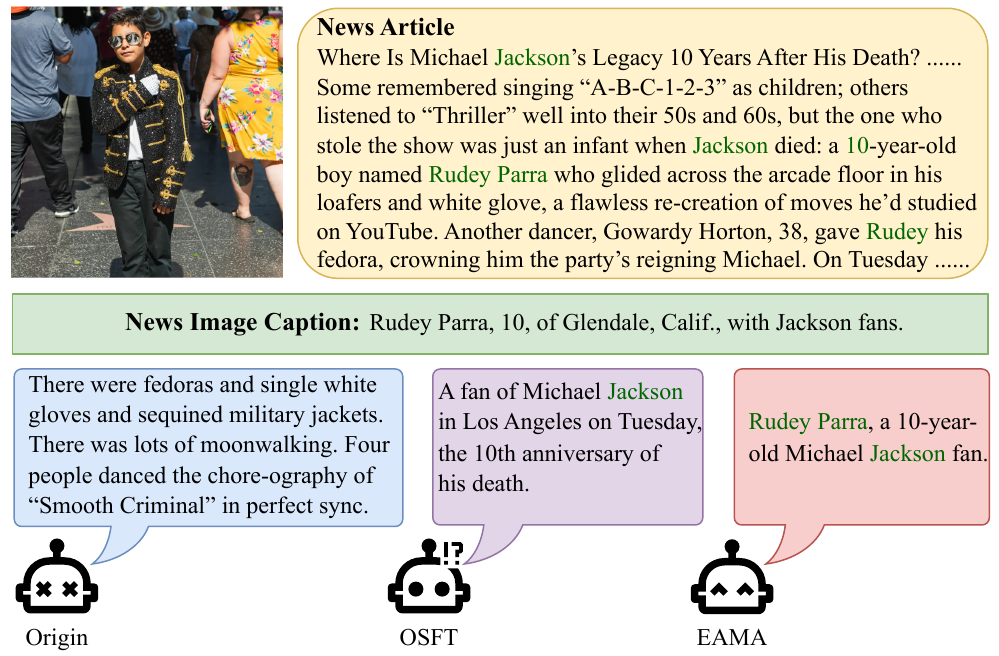}
    \caption{An example of news image caption task and responses of the MLLM in different settings. Origin denotes Zero\_Shot, OSFT denotes finetuning the MLLM in official settings. EAMA denotes our approach.}
    \label{fig:intro1}
\end{figure}

News image captioning, a variant of image captioning task that aims to generate the image caption which contains more specific information, has developed a lot as a standard multimodal task in recent years\citep{10.1145/3503161.3547883, rajakumar-kalarani-etal-2023-lets, qu-etal-2024-visually}.  In this task, models are required to generate an informative caption, rich in entities such as the names of people or news events, based on the provided news image and associated news article as illustrated in the example shown in Figure \ref{fig:intro1}. According to \citep{Tran_2020_CVPR}, over 96\% news image captions contain entity information and approximately 26\% words of news image captions are entities in common news image captioning datasets. Compared with general image captioning task, news image captioning task requires model to not only understand the overall information of images, but also apply the knowledge of model to extract related specific information from associated news articles.

Recently, multimodal large language models are developing rapidly and have shown promising capability to address traditional image captioning task. Therefore as a variant of image captioning task, it is natural to consider MLLMs as a promising technique for the news image captioning task. What's more, news image captioning task requires extracting related information from a long news article, which is exactly a task that can be handled by Large Language models because they contains extensive knowledge and exhibits powerful text processing ability. Since MLLMs mostly contain a LLM component, we believe MLLMs is potentially good at news image captioning task.

However, we find that current MLLMs such as InstructBLIP still face some challenges on news image captioning task: (1) Current MLLMs are not good at dealing with multimodal entity information under zero-shot setting. The ability of generating entities remains insufficient after simply finetuned on news image captioning dataset. An example is shown in Figure \ref{fig:intro1}. (2) While MLLMs are capable of processing long textual information and can take almost entire news articles as textual input, the generation of news image captions still requires a trade-off between sufficiency and conciseness of textual input information.

To address the problems we discovered, we propose EAMA, an Entity-Aware Multimodal Alignment based approach for News Image Captioning. Specifically, we first design two entity-aware alignment tasks, which are Entity-Aware Sentence Selection task and Entity Selection task. Then we align the MLLM on original news image captioning task together with two alignment tasks. Finally, we utilize the aligned MLLM to extract some relevant information from the news article to supplement the textual input context. Our method not only obtains the best result on all automatic metrics but also achieves the competitive performance on entity generation.

Our contributions are \footnote{Our code and data will be released to the community to facilitate future research.}:
\begin{itemize}

\item We find that current MLLMs' ability of dealing with multimodal entity information is insufficient under zero-shot setting and even after simply finetuned. Besides, generating high-quality news image captions with the MLLM still requires a trade-off between sufficiency and conciseness of textual input information.

\item We design two alignment tasks to promote the MLLM's ability for handling multimodal entity information. The aligned MLLM itself will explicitly extract the entity-related information to supplement textual input context while generating news image captions.



\item Our research conducts comprehensive experiments on two news captioning datasets, GoodNews and NYTimes800k, to demonstrate our analysis. Our proposed method EAMA outperforms all previous methods on both GoodNews and NYTimes800k Datasets.



\end{itemize}

\section{Related Works}


\subsection{News Image Captioning}
News image captioning\citep{Biten_2019_CVPR, Tran_2020_CVPR, liu-etal-2021-visual} is a variant of the general image captioning task which takes news images and associated news articles as input to generate informative captions. 
In early stage, models(\citet{lu-etal-2018-entity}, Avg+CtxIns\citep{Biten_2019_CVPR}) often generate templates, and then extract entities from news articles. Then, end-to-end methods\citep{Tran_2020_CVPR, yang-etal-2021-journalistic, 10.1145/3503161.3547883, qu-etal-2024-visually} generate the final caption in one step and achieve higher performance.
\citet{rajakumar-kalarani-etal-2023-lets} pretrains OFA model with extra data, and then finetunes OFA on news image captioning datasets. 
\citet{xu2024crossmodal} adapts finetuned CLIP to construct confidence-aware prompts to fine-tune LLaVA, and construct confidence-feedback prompts to further enhance its capabilities. \citet{10595419} proposes a rule-driven news captioning method, which can generate image descriptions following designated rule signal.
As for alignment, \citet{qu-etal-2024-visually} leverages a supervised face-name alignment method to refine face representations, but it only focuses on people's names. And its alignment method requires modifying model architecture and necessitates that the image contains a facial region. Our approach aligns the MLLM with newly designed alignment tasks focusing on general types of entities with data from the news image caption dataset without additional data.

Since news articles are long for previous models, many methods have been proposed. JoGANIC\citep{yang-etal-2021-journalistic} proposes MSTR to read more information. ICECAP\citep{10.1145/3394171.3413576} trains an VSE++ model to retrieve relevant sentences where entities are replaced with categories and then generate the template caption and predict the position of entity tokens in relevant sentences. \citet{zhou-etal-2022-focus} utilizes finetuned CLIP model and OpenNRE to obtain the key local context.
However, MSTR requires much longer input and most retrieval-based methods require an extra cross-modal retrieval module and an extra training cost while the caption model itself remained unimproved. \citet{xu2024crossmodal} also adapts CLIP as an image-text matching module. They construct confidence-aware prompts with CLIP to improve caption model LLaVA. However, they construct potential-positive training samples of CLIP by calculating the textual similarity between the input sentence and the caption. According to previous research \cite{zhou-etal-2022-focus}, captions often contain many non-visual entities, training the image-text matching module CLIP on these data may not be rigorous. The confidence-aware prompts generated by the trained CLIP are also not rigorous to train LLaVA. 
Our approach aligns the MLLM to promote its ability on handling multimodal entity information. The aligned MLLM can be used to explicitly extract entity-related information to supplement the textual input without extra modules. Our alignment tasks can help model focusing on both visually aware and non-visually aware entity information.
 

\subsection{MLLMs and Alignment}

Multimodal large language models(MLLMs) have developed rapidly in recent years. BLIP-2\citep{li2023blip} applys Q-Former architecture to transform image input into LLMs input tokens. LLaVA\citep{liu2023visual} and LLaVA-v1.5\citep{liu2023improved} utilize linear layers or perceptrons to map the vision features into the inputs of LLMs. Through instruction tuning on BLIP2, InstructBLIP\citep{dai2023instructblip} gains the ability to follow the instruction on different tasks. There are many other MLLMs such as mPLUG-Owl\citep{ye2023mplug}, Otter\citep{li2023otter} and Qwen-VL \citep{bai2023qwen}. Among all MLLMs, GPT-4V\citep{openai2023gpt4} is the most powerful one now. We select some of these MLLMs on our research. 

There are numerous studies on multimodal alignment tasks recent years. The most traditional task is image-text matching in CLIP\citep{pmlr-v139-radford21a}. There are also fine-grained alignment tasks such as Region-Tag and Region-Noun alignment tasks in \citet{Zhou_2022_CVPR}. MLLMs such as \citep{li2023blip, ye2023mplug} often align textual features and visual features on multimodal tasks with language modeling loss.

\section{Approach}
\subsection{Overview}



\begin{figure*}[tbp]
    \centering
    \small
    \includegraphics[width=\textwidth]{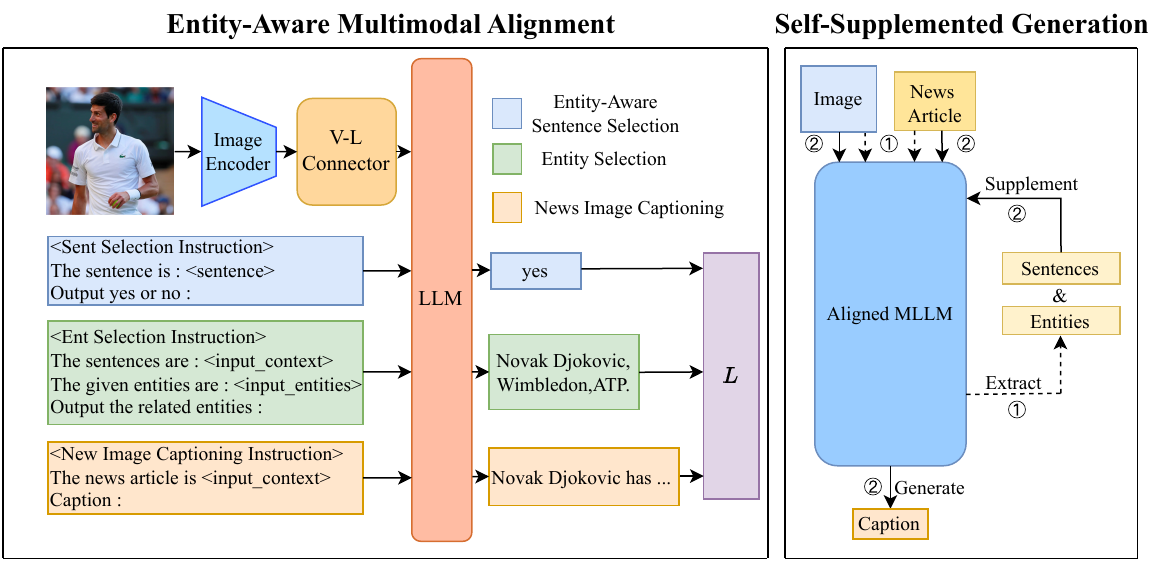}
    \caption{An overview of our approach: \textbf{EAMA}. The left part represents Entity-Aware Multimodal Alignment with Entity-Aware Sentence Selection, Entity Selection and News Image Captioning task. The right part represents the Self-Supplemented Generation for the news image caption given the news article and the news image. The aligned MLLM first extracts related sentences and entities and then generates the news image caption with extracted information as a supplement.}
    \label{fig:overview}
\end{figure*}

MLLMs (Multimodal Large Language Models) have been proved highly effective across diverse multimodal tasks, so we craft our methodology based on MLLMs. Figure \ref{fig:overview} offers a concise overview, delineating the alignment training and caption generation stages. Typically, MLLM architectures comprise an image encoder, a Vision-Language (V-L) Connector, and a large language model.

During our alignment training stage, we commence by simultaneously training the MLLM on three tasks: Entity-Aware Sentence Selection, Entity Selection, and News Image Captioning. This alignment aims to enhance the MLLM's capability to handle multimodal entity information.

We employ a language modeling loss as the optimization objective across all tasks. Typically, within a given dataset $D$, each data instance comprises three components: the input image $I$, the textual input context $T$, and the target output text $O=(o_{1},...,o_{n})$, with $o_{i}$ denoting a token. Thus, the general multimodal language modeling loss can be formulated as:
%


\begin{equation}
\centering 
\resizebox{1.02\hsize}{!}{
$\mathcal{L}_{\mathrm{LM}}([I;T;O]) = - \dfrac{1}{n} \sum\limits_{i=1}^{n}\log \mathbf{P}(o_i|o_1, ..., o_{i-1};I;T)$
}
\label{eq:language-modeling-loss}
\end{equation}
where the conditional probability $\mathbf{P}$ is calculated by neural network with multimodal features. 

For example, in News Image Captioning task, the input image $I$ is the news image, the textual input context $T$ is the news article, while the target output text $O$ is the given caption. Thus for a News Image Captioning dataset $D$, the corresponding News Image Captioning loss can be denoted as:

\begin{equation}
\centering 
\mathcal{L}_{\mathrm{LM}}^{\mathrm{CAP}} = \mathbb{E}_{([I;T;O])\sim D }\mathcal{L}_{\mathrm{LM}}([I;T;O]) 
\label{eq:language-modeling-loss-caption}
\end{equation}


In our caption generation stage, we leverage our aligned MLLM to extract key entities and entity-related sentences from news articles using multimodal input, without incurring additional modules. Subsequently, our aligned MLLM integrates this information to supplement the original textual input context and generate the news image caption.

\subsection{Entity-Aware Multimodal Alignment}
\noindent\textbf{Entity-Aware Sentence Selection}
Previous research frequently entails selecting sentences related to the caption from the provided news article \citep{zhou-etal-2022-focus}. A common approach involves employing an image-text matching model, such as CLIP, to extract relevant sentences based on news images. However, this method incurs additional training stages because CLIP is trained on open-domain datasets, which differ from captions enriched with entities on news image captioning datasets.

Interestingly, the image-text matching task is also commonly employed to align visual and textual features. Therefore, we aim to construct a specialized image-text matching task from news image captioning data to align vision features with entity-aware textual features. Moreover, considering the capability of MLLMs to follow instructions and handle multiple tasks, the process of selecting sentences related to the caption can be seamlessly integrated. Compared with previous methods, this integration not only streamlines the workflow but also reduces the need for separately training an image-text matching module.

Previous research \citep{zhou-etal-2022-focus} underscores the significance of entity information in news image captioning tasks. As a result, general image-text matching tasks may not be well-suited for entity-aware tasks like news image captioning. Hence, we introduce an Entity-Aware Sentence Selection task as one of our Entity-Aware Multimodal Alignment tasks.

To construct Entity-Aware Sentence Selection data, we leverage sentences from the news article as the text source. For a given instance comprising the news image $I$, the news article $T={s_1, s_2, ..., s_m}$ consisting of $m$ sentences, and the news image caption $C$, we first identify the visual entities $E = (e_1, e_2, ..., e_{n})$ mentioned in the caption. Notably, the visual entities exclude some entities invisible in the news image, such as `DATE'.

Subsequently, we proceed to select sentences from the news article that contain the same visual entities as those mentioned in the caption. These selected sentences, along with the corresponding caption itself, are regarded as positive samples. Conversely, negative samples are derived from the remaining sentences of the article.

Formally, for Entity-Aware Sentence Selection task, the input image $I$ is the news image, the input context is one sentence $s_{i}$, and the corresponding output is given as:
\begin{equation}
    O = \left\{ \begin{aligned}&yes,\ \exists e_{j} \in E, s.t. e_{j} \ \in  s_{i} \\& no,\ otherwise \end{aligned} \right.
\end{equation}


For the Entity-Aware Sentence Selection task, each group consists of sentences containing the most visual entities mentioned in the caption, along with the target caption, serving as positive samples. Additionally, sentences are chosen from those devoid of any visual entity mentioned in the caption, serving as negative samples.

On dataset $D$, the final loss of Entity-Aware Sentence Selection task is
\begin{equation}
    \mathcal{L}^{\mathrm{SENT}}_{\mathrm{LM}} = \mathbb{E}_{([I;s_i;O]) \sim D} \mathcal{L}_{\mathrm{LM}}([I;s_{i};O])
\end{equation}

Unlike traditional image-text matching methods, which often randomly select captions corresponding to other images as negative samples, our approach extracts sentences from the same article that do not contain the visual entities as hard negative samples. These sentences frequently present a more similar distribution and cover the same topic, only lacking visually related entity information. Model can align the multimodal entity information during the alignment task.

\noindent\textbf{Entity Selection}
In contrast to the Entity-Aware Sentence Selection task, Entity Selection entails a more fine-grained alignment process. Within the Entity Selection task, we identify entities present in both the input text and the caption, preserving the order of appearance in the caption as the output sequence. The model is fed with the news image, its associated input context, and recognized entities from the input context as input, predicting the output entities accordingly. The objective of this task is to enhance the model's proficiency in recognizing entities relevant to captions, a crucial aspect for generating entity-rich captions.

Specifically, for each instance of news image captioning dataset, we recognize the entities $E_{T}$ from news article $T$ and $E$ from the news image caption. The output is 

\begin{equation}
O_e = (e_i, e_i \in E \cap E_{T})
\end{equation}

We combine the news article $T$ and recognized entities $E_{T}$ into the final input context $T^{*}$.
In dataset $D$, the final loss of Entity Selection task is
\begin{equation}
    \mathcal{L}^{\mathrm{ENT}}_{\mathrm{LM}} = \mathbb{E}_{([I;T^{*};O_e]) \sim D} \mathcal{L}_{\mathrm{LM}}([I;T^{*};O_e])
\end{equation}

It is noteworthy that, unlike the Entity-Aware Sentence Selection task, the Entity Selection task encompasses both visual and non-visual entities, a concept inspired by \citep{zhou-etal-2022-focus}. Given that models can access both the news image information and the background context information within the news article, we contend that they possess sufficient contextual cues to identify non-visual entities. These non-visual entities are frequently associated with the visual entities in the news article and play a pivotal role in caption generation.


\begin{figure*}
    \centering
    \includegraphics[width=\textwidth]{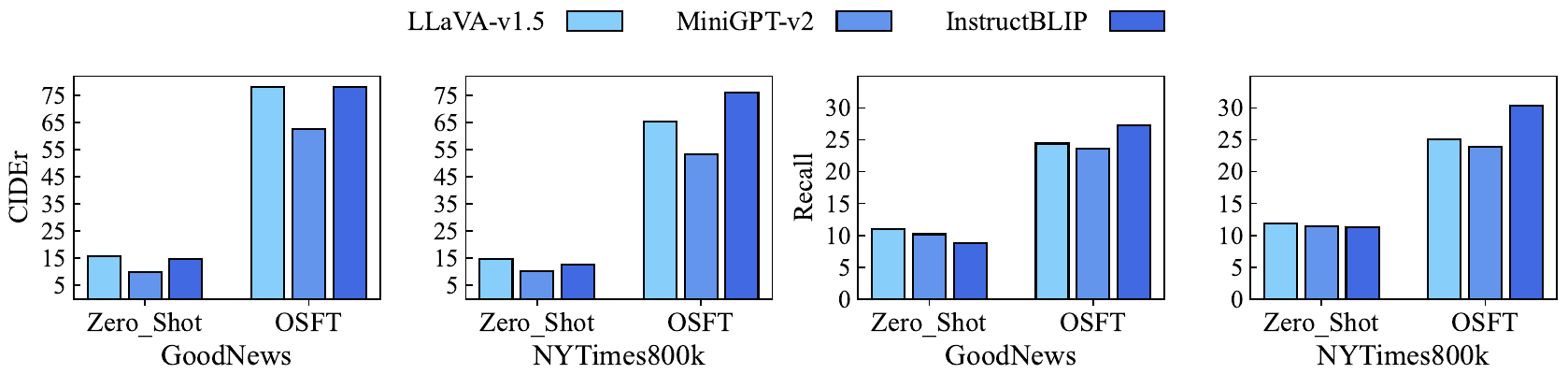}
\caption{
Performance of MLLMs in Zero\_Shot and Official supported Supervised Finetune (OSFT) setting. Following the OSFT settings, we train the V-L Connector in MiniGPT-v2 and InstructBLIP, and fully train the V-L Connector together with LLM in LLaVA-v1.5. LLaVA-v1.5, MiniGPT-v2 and InstructBLIP are implemented with 7B checkpoints.}
\label{fig:mllm_performance}
\vspace{-1.5em}
\end{figure*}

\subsection{Self-Supplemented Generation}
To enhance news image caption generation, models rely on concise and pertinent textual information. Previous research often employs a specific length limitation from the start of news articles or paragraphs surrounding the provided image due to token limitations. In our method, the aligned MLLM initially extracts related entities and sentences which contain entities relevant to the given news image from the entire news article. Because the aligned MLLM is trained on the Entity-Aware Sentence Selection task and Entity Selection task during the alignment training stage, this extraction stage doesn't incur additional modules.

Once we acquire related entities and entity-related sentences, we seamlessly integrate them with the default textual input context. To ensure coherence, we eliminate duplicated sentences and combine the rest sentences into the final textual input context, maintaining their original order in the news article. To prevent the final textual input context from becoming overly long, we restrict the overall length of the final input context. More details can be seen in Appendix \ref{sec:implementation-details}.

\begin{table*}[ht]
\centering
\setlength{\tabcolsep}{2mm}{
\begin{tabular}{m{1em}  c  c c c c c c} 
\bottomrule %
   & \multirow{2}*{Method}  & \multirow{2}*{BLEU-4} & \multirow{2}*{METEOR} & \multirow{2}*{ROUGE} & \multirow{2}*{CIDEr} & \multicolumn{2}{c}{Named entities}  \\
   &  & &  &  &  & P & R  \\
 \hline
 \multirow{10}*{\rotatebox{90}{GoodNews}} & Avg+CtxIns\citep{Biten_2019_CVPR} 
      & 0.89 & 4.37 & 12.20 & 13.10 & 8.23 & 6.06  \\ 
      & ICECAP\citep{10.1145/3394171.3413576} & 1.96 & 6.01 & 15.70 & 26.08 & / & / \\

  & Tell\citep{Tran_2020_CVPR}   & 6.05 & 10.30 & 21.40 & 53.80 & 22.20 & 18.70  \\ 
  & JoGANIC\citep{yang-etal-2021-journalistic}  & 6.83 & 11.25 & 23.05 & 61.22 & 26.87 & 22.05  \\
  & NewsMEP\citet{10.1145/3503161.3547883} & 8.30 & 12.23 & 23.17 & 63.99 & 23.43 & 23.24 \\
  & Rule-driven\citep{10595419} & 8.18 & 12.50 & 23.56 & 71.58 & 25.51 & 23.68 \\
  & \citet{rajakumar-kalarani-etal-2023-lets} & 7.14 & 11.21 & 24.30 & 72.33 & 24.37 & 20.09 \\
  & \citet{qu-etal-2024-visually}   & 8.60 & 12.39 & 23.38 & 71.96  & 24.30 & 25.54 \\ 
  & \citet{xu2024crossmodal} & 8.49 & 12.88 & 26.22 &  83.52 & \textbf{30.19} & 26.57      \\ \cline{2-8}
  & InstructBLIP (OSFT) & 9.53 & 13.54 & 25.61 & 78.03 & 25.89 & 27.33 \\
  

  & \textbf{EAMA} & \textbf{10.04} & \textbf{13.95} & \textbf{27.06} & \textbf{87.70} & 27.58 & \textbf{28.92} \\

  \cline{2-7}
 \hline 
  \multirow{8}*{\rotatebox{90}{NYTimes800k}} 
  & Tell\citep{Tran_2020_CVPR}   & 6.30 & 10.30 & 21.70 & 54.40 & 24.60 & 22.20  \\ 
  & JoGANIC\citep{yang-etal-2021-journalistic}  & 6.79 & 10.93 & 22.80 & 59.42 & 28.63 & 24.49  \\
  & NewsMEP\citep{10.1145/3503161.3547883} & 9.57 & 13.02 & 23.62 & 65.85 & 26.61 & 28.57 \\
  & Rule-driven\citep{10595419} & 9.41 & 13.10 & 24.42 & 72.29 & 28.15 & 28.80 \\
  & \citet{rajakumar-kalarani-etal-2023-lets} & 7.54 & 11.27 & 23.28 & 66.41 & 28.11 & 23.25 \\
  & \citet{qu-etal-2024-visually}   & 9.24 & 12.57 & 23.44 & 71.65 & 26.88 & 28.59 \\ 
  & \citet{xu2024crossmodal} & 9.07 & 13.17 & 26.48 & 83.72 & \textbf{32.38} & 30.08 \\ \cline{2-8}
  & InstructBLIP (OSFT) & 10.05 & 13.63 & 25.45 & 75.95 & 27.39 & 30.37 \\
  & \textbf{EAMA} & \textbf{11.03} & \textbf{14.22} & \textbf{27.15} & \textbf{87.00} & 29.79 & \textbf{32.24} \\ 

\bottomrule
\end{tabular}}
\caption{Experimental results on GoodNews and NYTimes800k datasets compared with other models. P and R denote Precision and Recall of generated Named entities. InstructBLIP(OSFT) refers to InstructBLIP trained with official scripts on 7B checkpoints. We directly cite the results of models which are not MLLMs from researches by \citet{rajakumar-kalarani-etal-2023-lets, qu-etal-2024-visually}.}
\label{table:Overall Score}
\vspace{-1.0em}
\end{table*}

\begin{table*}[ht]
\renewcommand{\arraystretch}{1.2}
\centering
\setlength{\tabcolsep}{3mm}{
\begin{tabular}{c c c c c c c} 
\bottomrule %
\multirow{2}*{Method} & \multirow{2}*{BLEU-4} & \multirow{2}*{METEOR} & \multirow{2}*{ROUGE} & \multirow{2}*{CIDEr} & \multicolumn{2}{c}{Named entities}  \\
  & & &  &  & P & R  \\  \hline
InstructBLIP                         & 10.05 & 13.63 & 25.45 & 75.95 & 27.39 & 30.37   \\ 
Base                  & 10.88 & 14.09 & 26.60 & 82.70 & 29.22 & 31.32 \\ 
Align(SENT $+$ CAP)      & 10.92 & 14.08 & 26.68 & 83.60 & 29.42 & 31.45\\
Align(ENT $+$ CAP)                  & 10.97 & 14.20 & 26.87 & 84.28 & 29.26 & 31.58\\
Align(SENT $+$ ENT $+$ CAP)  & 10.80 & 14.08 & 26.80 & 84.45 & 29.15 & 31.45\\

\textbf{EAMA}                              & \textbf{11.03} & \textbf{14.22} & \textbf{27.15} & \textbf{87.00} & \textbf{29.79} & \textbf{32.24} \\


\bottomrule
\end{tabular}}
\caption{Ablation study results on the NYTimes800k dataset. 
InstructBLIP refers to OSFT InstructBLIP. Base indicates fully finetuning InstructBLIP on the news image caption task. Align refers to aligning InstructBLIP with different alignment tasks. EAMA denotes our entire approach. According to Sec \ref{subsec:data_metrics}, CIDEr is the most crucial metric because it places greater emphasis on uncommon words, which often include specific named entities.}
\label{table:ablation}
\vspace{-1.0em}
\end{table*}


\section{Experiment Setup}
\subsection{Datasets \& Metrics}
\label{subsec:data_metrics}
In our experiments, we utilize two publicly available large-scale News Image Captioning datasets\footnote{We don't have the official test split of VisualNews dataset, so we conduct experiments on the same two datasets with \cite{qu-etal-2024-visually} and \cite{xu2024crossmodal}.}: GoodNews \citep{Biten_2019_CVPR}, comprising 421K, 18K, and 23K samples for train, validation, and test splits respectively, and NYTimes800k \citep{Tran_2020_CVPR}, containing 763K, 8K, and 22K samples for train, validation, and test splits.

We follow the validation/test data split as previous researches and randomly select 100k samples for both train datasets. The average lengths of news articles and news image captions in the GoodNews dataset are 451 and 18 respectively. In NYTimes800k, the average lengths of articles and captions are 974 and 18 respectively.

We use BLEU-4 \citep{papineni2002bleu}, METEOR \citep{denkowski-lavie-2014-meteor}, ROUGE \citep{lin2004rouge} and CIDEr \citep{vedantam2015cider} as automatic evaluation metrics, following previous research \citep{Tran_2020_CVPR} \footnote{These metrics are calculated with the COCO caption evaluation toolkit\ \url{https://github.com/tylin/coco-caption}}. According to previous researches \citep{Tran_2020_CVPR, zhou-etal-2022-focus, zhang-wan-2023-exploring}, CIDEr is the most crucial metric because it places greater emphasis on uncommon words, which often include specific named entities. Additionally, we evaluate the Precision and Recall of named entities. To accomplish this, we employ SpaCy to recognize entities in both generated captions and gold captions and count the number of exact matches.

\subsection{Implementation Details}


When simply finetuning MiniGPT-v2, InstructBLIP, and LLaVA-v1.5, adhering to the official scripts, we fine-tune the linear projection layer of MiniGPT-v2, the Q-Former, and Fully Connected Layer of InstructBLIP, and both the MLP layer and LLM of LLaVA-v1.5 on checkpoints of 7B. Our training and inference prompts can be found in Appendix \ref{sec:training-inference-prompts}, which remain unchanged among our experiments. All training hyper-parameters follow their default configurations: the initial learning rates for InstructBLIP and LLaVA-v1.5 are set to $2\times10^{-5}$, while the initial learning rate for MiniGPT-v2 is set to $1\times10^{-5}$. During inference, we use a beam size of 5 in MLLMs. Mixed precision training with bp16 is employed to accelerate the training process. Our model takes approximately 2-3 days to train on four Nvidia A40 GPUs. More implementation details can be found in Appendix \ref{sec:implementation-details}.


\section{Experimental Results}
\subsection{Performance of MLLMs on NIC}
%
LLaVA-v1.5\citep{liu2023improved}, MiniGPT-v2\citep{chen2023minigptv2} and InstructBLIP\citep{dai2023instructblip} are three common MLLMs. We show the representative results in Figure \ref{fig:mllm_performance} and full evaluation results are shown in Table \ref{table-app:mllm_performance} in Appendix \ref{sec:app-perf-mllm}.

Firstly, the results under zero-shot setting confirm our initial hypothesis that MLLMs trained on general multimodal data struggle to generate informative news image captions. Notably, the Recall of named entities is lower than the Precision of named entities on MLLMs, especially for LLaVA-v1.5, which means MLLMs tend not to generate entities under zero-shot setting. In fact, the average number of generated entities (whether correct or not) by LLaVA-v1.5 are 1.22 and 1.24 on GoodNews and NYTimes800k datasets, while the average number of entities in references are 3.43 and 3.19 on these two datasets respectively. This substantial disparity lends further support to our observations.

Subsequently, we conduct straightforward finetuning of these three MLLMs with their official scripts on news image captioning dataset. Following their official scripts, we finetune the V-L Connector and LLM component of LLaVA-v1.5 and the V-L Connector component of MiniGPT-v2 and InstructBLIP. We refer to this training setting as OSFT. According to the results in Figure \ref{fig:mllm_performance}, this simple finetuning approach yields promising performance improvements on both the GoodNews and NYTimes800k datasets. Thus we believe that MLLMs are capable of multimodal Entity-Aware tasks when trained appropriately on multimodal data rich in entity information. Besides, we observe that the OSFT InstructBLIP obtains better performance than LLaVA-v1.5 and MiniGPT-v2 on both dataset, especially on NYTimes800k which contains longer news articles. We attribute this to the V-L Connector module of InstructBLIP, which compresses visual information into 32 tokens.  This enables InstructBLIP to accommodate more textual input compared to other MLLMs, making InstructBLIP the most suitable MLLM among all three MLLMs for news image captioning. Therefore, we select InstructBLIP for further investigation and experimentation. 

\subsection{Comparison Results of Different Methods}
\label{sec:Comparison_res}

Table \ref{table:Overall Score} presents a comparison of the performance of our proposed approach (EAMA) and the baseline models, including the OSFT InstructBLIP settings on two news image captioning datasets. The results in Table \ref{table:Overall Score} show that our method outperforms the OSFT InstructBLIP approach by 9.67 and 11.05 points in CIDEr score on the GoodNews and NYTimes800k datasets, respectively. Furthermore, our method EAMA achieves the best performance across all automatic metrics. Compared to the previous state-of-the-art (SOTA) method (\citet{xu2024crossmodal} for GoodNews and NYTimes800k), our model gains 4.18 and 3.28 points in CIDEr score on the GoodNews and NYTimes800k datasets, respectively. In addition to the superior performance on automatic evaluation metrics, our method also achieves the highest recall of named entities simultaneously. We also provide analysis on entity generation in Appendix \ref{sec:ana-ent-gen} and a case study in Appendix \ref{sec:app-case} to offer a comprehensive understanding of EAMA.

\subsection{Ablation Study}
We conduct ablation experiments on the NYTimes800k dataset to study the impact of different components of our approach. In addition to the OSFT of InstructBLIP, we also fully fine-tuned both the Vision-Language Connector and the Large Language Model, which we refer to as Base. The results in Table \ref{table:ablation} show that fully fine-tuning the Large Language Model is crucial for InstructBLIP to generate high-quality news image captions. We hypothesize that handling entity information using multimodal information is very challenging for its distinction from other common multimodal data, making fine-tuning the Vision-Language Connector alone insufficient.

We also conducted ablation studies to demonstrate the effectiveness of the Entity-Aware Sentence Selection and Entity Selection tasks individually. In Table \ref{table:ablation}, Align(SENT $+$ ENT $+$ CAP) means aligning the model on the Entity-Aware Sentence Selection, Entity Selection, and News Image Captioning tasks simultaneously. Align(SENT $+$ CAP) and Align(ENT $+$ CAP) follow the same pattern. From the ablation study results, we observe that both alignment tasks are effective when used separately. Both alignment tasks are effective and necessary components of our alignment framework.


EAMA refers to our approach, which first aligns the InstructBLIP model with all three tasks, and then applies Self-Supplemented Generation during the inference stage when generating news image captions. With the addition of Self-Supplemented Generation, EAMA further improves the performance from a CIDEr score of 84.45 to 87.00, and also achieves better generation of entities.

\subsection{Analysis on Textual Input}
\label{sec:textual_input_ana}
We would also like to investigate the influence of different textual input on news image captioning task. The full results are shown in Table \ref{table:analysis on Textual Input} in Appendix \ref{sec:app-ana-text}. In our experiments on the textual input for the news image captioning task, we observe that most previous methods limit the length of textual context due to the maximum token constraint. As indicated in \citep{Tran_2020_CVPR}, the average length of news articles in the NYTimes800k dataset is 974 words. Given that InstructBLIP has a maximum of 2048 tokens, with only 32 of them being visual tokens, we attempt to use entire news articles as input during both training and inference stages. This approach aims to leverage more textual information into the MLLM, potentially enhancing performance. However, our experiments reveal that the performance only increases marginally, from 82.70 to 83.17 in CIDEr score. This modest improvement is not satisfactory, considering that using such long textual input nearly doubles the time and memory costs during training and inference stages.

To research deeper, we conduct an additional experiment by testing the model with oracle-selected sentences as textual input. These oracle sentences averagely contain only 14\% of the words from entire news articles. They are chosen from news articles and contain the same entities as reference captions. We also test the model with oracle-selected entities which occur both in the news article and the caption. To obtain the oracle upper bound, We test the model with input textual context supplemented with both oracle-selected sentences and entities. While these scenarios are ideal and impossible in practice since we cannot access reference captions during inference, the results indicate that the oracle supplemented textual input can achieve competitive performance. Therefore, compared to simply expanding the length of textual input, we believe that selecting related entities and entity-related sentences for supplementation is more effective. 


The results further suggest that our supplemented textual input contains concise and relevant information for the news image captioning task, yielding better results. To ascertain that this improvement stems from self-supplementation rather than merely expanding input length, we expanded the original input to match the length of the supplemented input, following the traditional setting (labeled as Origin(Longer)). The results indicate that our self-supplemented generation doesn't solely improve performance by increasing input information. Instead, it relies on concise and entity-related input context.

\section{Conclusion}
In this work, we reveal that MLLMs trained on conventional multimodal data struggle with generating entities in news image captioning task. Simply fine-tuning MLLMs proves insufficient for enabling them to effectively handle multimodal entity information. To tackle this challenge, we introduce EAMA, which aligns the MLLM with two entity-aware alignment tasks. Subsequently, we utilize the aligned MLLM to supplement the textual input context, as we observe that generating high-quality news image captions necessitates entity-related and concise textual input context. Our proposed method, EAMA, not only achieves superior performance on automatic evaluation metrics but also obtains the competitive result of entity generation in news image captioning task. 

\section{Limitations}
We conduct our experiments mainly on three common MLLMs. The conclusion will be more general if we test more MLLMs. Our method requires big GPU memory both on training and testing stages, and is slower than previous models. With the Optimization technique such as flash-attention provided, the speed of our method will become faster. The data of our alignment tasks are extracted only from news image captioning tasks. The knowledge of news is updating rapidly. Our method still need to align the model on data which contains new updated knowledge to obtain the best performance.

\section{Ethics Statement}
The generation of our method is highly related to the pretrained multimodal checkpoints and the alignment task data. They may both introduce bias and hallucination into the generated captions which may cause inaccuracy news report. We recommend to manual check the results before actually using in the real situation.

\bibliography{custom}

\appendix

\section{Training and Inference Prompts}
\label{sec:training-inference-prompts}
\begin{figure*}[tbp]
    \centering
    \includegraphics[width=0.97\textwidth]{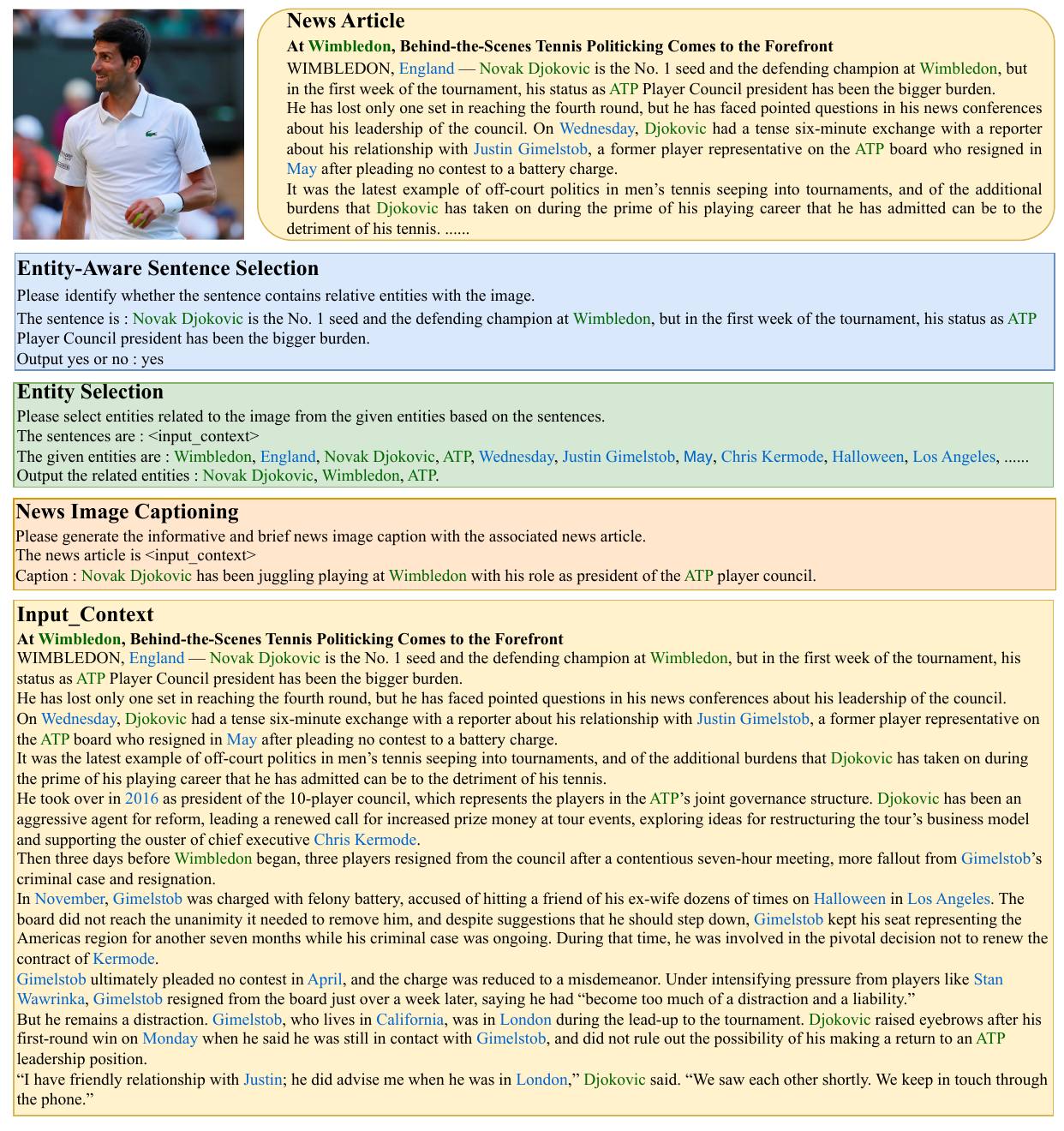}
    \caption{An example of our prompts for Entity-Aware Sentence Selection, Entity Selection and News Image Captioning tasks. We label entities in the caption in \textcolor[RGB]{0,150,0}{green}. We label entities in the input context but not in the caption in \textcolor[RGB]{0,66,204}{blue}.}
    \label{fig:example_prompt}
\end{figure*}

We show our prompts in Figure \ref{fig:example_prompt} for our Entity-Aware Sentence Selection, Entity Selection and News Image Captioning tasks. We use the same prompt during training and inference stages.

\begin{table*}[htp]
\centering 
\setlength{\tabcolsep}{1.8mm}{
\begin{tabular}{c c c c c c c c c} 
\bottomrule[1.2pt]
\multirow{2}*{Dataset} & \multirow{2}*{MLLM} & \multirow{2}*{Setting} & \multirow{2}*{BLEU-4} & \multirow{2}*{METEOR} & \multirow{2}*{ROUGE} & \multirow{2}*{CIDEr} & \multicolumn{2}{c}{Named entities}\\
 & & & & & & & P& R \\
  \hline
  \multirow{6}*{\rotatebox{90}{GoodNews}} 
                          & \multirow{2}*{LLaVA-v1.5}   & Zero\_Shot & 2.63 & 8.72 & 14.03 & 15.85 & 17.19 & 10.96 \\
                          &                             & OSFT       & 8.48 & 12.37 & 25.22 & \textbf{78.03} & 25.69 & 24.41 \\ \cline{2-9}
                          & \multirow{2}*{MiniGPT-v2}    & Zero\_Shot & 1.81 & 7.62 & 11.81 & 9.77 & 15.19 & 10.17  \\
                          &                             & OSFT       & 8.05 & 12.22 & 22.78 & 62.42 & 21.32 & 23.62 \\ \cline{2-9}

                          & \multirow{2}*{InstructBLIP} & Zero\_Shot & 2.52 & 6.68 & 10.78 & 14.82 & 13.34 & 8.78  \\
                          &                             & OSFT       & \textbf{9.53} & \textbf{13.54} & \textbf{25.61} & \textbf{78.03} & \textbf{25.89} & \textbf{27.33}  \\
  \midrule[0.8pt]
  \multirow{6}*{\rotatebox{90}{NYTimes800k}}
                          & \multirow{2}*{LLaVA-v1.5} & Zero\_Shot   & 2.65 & 8.75 & 13.53 & 14.66 & 17.24 & 11.82 \\
                          &                           & OSFT         & 7.94 & 11.67 & 23.48 & 65.28 & 25.92 & 25.01  \\ \cline{2-9}

                          & \multirow{2}*{MiniGPT-v2}    & Zero\_Shot & 1.85 & 7.69 & 11.56 & 10.21 & 14.15 & 11.41 \\ 
                          &                             & OSFT       & 7.38 & 11.32 & 21.27 & 53.30 & 21.25 & 23.89 \\ \cline{2-9}

                          & \multirow{2}*{InstructBLIP} & Zero\_Shot & 2.52 & 7.41 & 11.01 & 12.74 & 12.79 & 11.21  \\
                          &                             & OSFT & \textbf{10.05} & \textbf{13.63} & \textbf{25.45} & \textbf{75.95} & \textbf{27.39} & \textbf{30.37}  \\
\bottomrule[1.2pt]
\end{tabular}}
\caption{
Performance of MLLMs in Zero\_Shot and Official supported Supervised Finetune (OSFT) setting. Following the OSFT settings, we train the V-L Connector in MiniGPT-v2 and InstructBLIP, and fully train the V-L Connector together with LLM in LLaVA-v1.5. LLaVA-v1.5, MiniGPT-v2 and InstructBLIP are implemented with 7B checkpoints.}
\label{table-app:mllm_performance}

\end{table*}

\section{Implementation details}
\label{sec:implementation-details}
We restrict the max length of multimodal input tokens to 1024 and the max length of output new tokens to 50 on most experiments. We set the max length of multimodal input tokens to 2048 while taking the full news article as textual input for the specific experiment in Section \ref{sec:textual_input_ana}. Following previous research\citep{Tran_2020_CVPR}, the traditional setting to obtain textual input on the GoodNews dataset involves extracting the first 500 words of the entire news article. Conversely, on the NYTimes800k dataset, the traditional textual input(as loc\_context) entails obtaining words around the image, as the NYTimes800k dataset provides information about the image position within the entire article. Our method extracts sentences which contain entities relevant to the given news image from the entire news article first, then applies these sentences to supplement the textual input context. We set a maximum limitation of 600 words during this stage. Our method will then extract the related entities from the news article and combine them with prompt such as "The possible related entities are:" into textual input context. In Entity-Aware Multimodal Alignment stage, we organize the data into mini-groups, each containing 2 instances of the News Image Captioning task, 1 set of instances of the Entity-Aware Sentence Selection task, and 1 instance of the Entity Selection task. Following previous researches, we employ SpaCy to recognize entities during our research. 

In the OSFT setting, we utilize the official scripts for InstructBLIP, MiniGPT-v2, and LLaVA-v1.5. Since there are no official scripts available for fully finetuning InstructBLIP, we re-implement InstructBLIP to enable fully finetuning with deepspeed as Base in our experiments. 

When aligning our model with the two Entity-Aware alignment tasks, the final loss is the weighted sum of loss in there tasks. We set the weight of $\mathcal{L}^{\mathrm{SENT}}_{\mathrm{LM}}$, $\mathcal{L}^{\mathrm{ENT}}_{\mathrm{LM}}$, and $\mathcal{L}_{\mathrm{LM}}^{\mathrm{CAP}}$ as follows: 0.5, 0.25, and 1 respectively for the GoodNews dataset, and 0.25, 0.75, and 1 respectively for the NYTimes800k dataset. These values are determined based on their performance on the validation dataset.

\section{Performance of Different MLLMs}
\label{sec:app-perf-mllm}
We present the whole evaluation results of different MLLMs in Table \ref{table-app:mllm_performance}. The results correspond with our observation that MLLMs perform badly under zero\_shot settings and simple finetuning yields promising results, revealing the potential of MLLMs on News Image Captioning task.

\begin{table*}[ht]
\centering
\setlength{\tabcolsep}{1.3mm}{
\begin{tabular}{c c c c c c c c c} 
\bottomrule %
\multirow{2}*{Method} & Textual Input & Textual Input & \multirow{2}*{BLEU-4} & \multirow{2}*{METEOR} & \multirow{2}*{ROUGE} & \multirow{2}*{CIDEr}  & \multicolumn{2}{c}{Named entities}  \\
  & NIC Training & NIC Inference & & & &  & P & R  \\  \hline
Base & Origin & Origin                       & 10.88 & 14.09 & 26.60 & 82.70 & 29.22 & 31.32 \\
Base  & Full & Full                         & 10.76 & 13.84 & 26.18 & 83.17 & 27.71 & 30.03 \\
Base  & Origin & Origin(Longer)         & 10.60 & 13.80 & 26.28 & 81.75 & 28.95 & 30.57 \\ 
Base  & Origin & Oracle(SENT)                       & 11.24 & 14.50 & 27.70 & 87.56 & 33.42 & 34.80\\ 
Base  & Origin & Oracle(ENT)                       & 11.63 & 14.76 & 27.90 & 90.44 & 32.04 & 34.24 \\ 
Base  & Origin & Oracle(SENT+ENT)                  & \textbf{11.89} & \textbf{15.07} & \textbf{28.79} & \textbf{94.27} & \textbf{35.71} & \textbf{37.10} \\ 
\hline 
Align  & Origin & Origin                 & 10.80 & 14.08 & 26.80 & 84.45 & 29.15 & 31.45\\
Align  & Origin & Origin(Longer)   & 10.75 & 14.00 & 26.70 & 83.50 & 29.25 & 31.16\\
\textbf{EAMA}   & Origin & Supplemented                  & \textbf{11.03} & \textbf{14.22} & \textbf{27.15} & \textbf{87.00} & \textbf{29.79} & \textbf{32.24} \\

\bottomrule
\end{tabular}}
\caption{Analysis of Textual Input in News Image Captioning (NIC) on NYTimes800k dataset. Base indicates fully finetuning InstructBLIP on the NIC task. Align means align InstructBLIP with all three alignment tasks. Origin denotes the traditional textual input of NIC. Full denotes the entire news article. Supplemented denotes fusing the extracted sentences and entities with Origin under a limited length. Origin(Longer) denotes obtaining textual input of the same maximum length as Supplemented but using traditional setting. Oracle(SENT) denotes using the sentences containing the same entities as captions and Oracle(ENT) denotes using entities exactly both appearing in the new article and caption. Oracle(SENT+ENT) denotes using the oracle sentences and entities to supplement the input textual context, serving as an ideal upper bound.}
\label{table:analysis on Textual Input}
\end{table*}


\section{Result of Textual Input Experiment}
\label{sec:app-ana-text}
We present the full evaluation results in Table \ref{table:analysis on Textual Input}.

\section{Analysis of Entity Generation}
\label{sec:ana-ent-gen}

\begin{table}[htbp]
\centering 
\setlength{\tabcolsep}{5mm}{
\begin{tabular}{c c c c}
\bottomrule
    \multirow{2}*{Method}  & \multicolumn{2}{c}{Named entities} \\                                   & In    & Out \\
    \hline 
    Tell*          & 38.22 & 4.56  \\
    JoGANIC*       & 38.83 & 4.88 \\
    InstructBLIP   & 48.81 & 11.23 \\ 
    \textbf{EAMA}           & \textbf{50.64} & \textbf{13.16} \\
\bottomrule
\end{tabular}
}
\caption{The Recall of Entities in and out of textual input context on NYTimes800k dataset. All methods except EAMA take loc\_context as textual input, EAMA takes supplemented textual input. We cite the results of Tell and JoGANIC from previous researches\cite{zhang-wan-2023-exploring}.}
\label{table:entities explore}
\end{table}

\noindent \textbf{In-Out Textual Input Entity Generation} 
According to prior research\cite{zhang-wan-2023-exploring}, evaluating the generation of entities that appear in both the textual input context and captions can provide better insight into the models' ability on handling entity information. Therefore, we further analyze the Recall of entities in and outside of the textual input context.

In news image captioning task, the Recall of entities in the textual input context is more crucial. This is because generating entities that are not present in the textual input context not only poses a greater challenge but also risks introducing more hallucinations or spurious information into the generated captions. Therefore, achieving high Recall of entities in the textual input context is an essential aspect of evaluating the performance of models in news image captioning tasks.

All results are listed in Table \ref{table:entities explore}, InstructBLIP obtain high Recall especially for entities of caption occurring in the textual input context. Our method obtains the highest Recall of entities appearing both in and outside of textual input context, which is consistency with the our previous results.

\begin{table}[ht]
\centering 
\setlength{\tabcolsep}{5mm}{
\begin{tabular}{c c c c}
\bottomrule
    \multirow{2}*{Method}  & \multicolumn{2}{c}{Named entities} \\
                           & P       & R \\ \hline 
    \textbf{EAMA}                   & 29.70   & 28.21 \\
\bottomrule
\end{tabular}}
\caption{The Precision and Recall of Entities which are out of train dataset in our approach: EAMA, on NYTimes800k dataset.}
\label{table:out-train entities}

\end{table}

\noindent \textbf{Out-of-Train Entity Generation}
To assess our method's capability to predict entities not present in the training data, we examine captions in the NYTimes800k test dataset, focusing on entities that do not appear in the training data. The entire test set comprises 70,002 entities in the reference captions, with 13,481 of them not appearing in the textual input contexts or captions of the training data. Our approach, EAMA, successfully predicts 3,803 entities in the generated captions, resulting in a Recall of 28.21. The Precision of the out-of-train entities is 29.70, as shown in Table \ref{table:out-train entities}. Both the Recall and Precision metrics are comparable to the results for all entities presented in Section \ref{sec:Comparison_res}, demonstrating the generalization ability of EAMA for generating out-of-train entities.


\section{Case Study}
\label{sec:app-case}
\begin{figure*}[tbp]
    \centering
    \includegraphics[width=0.97\textwidth]{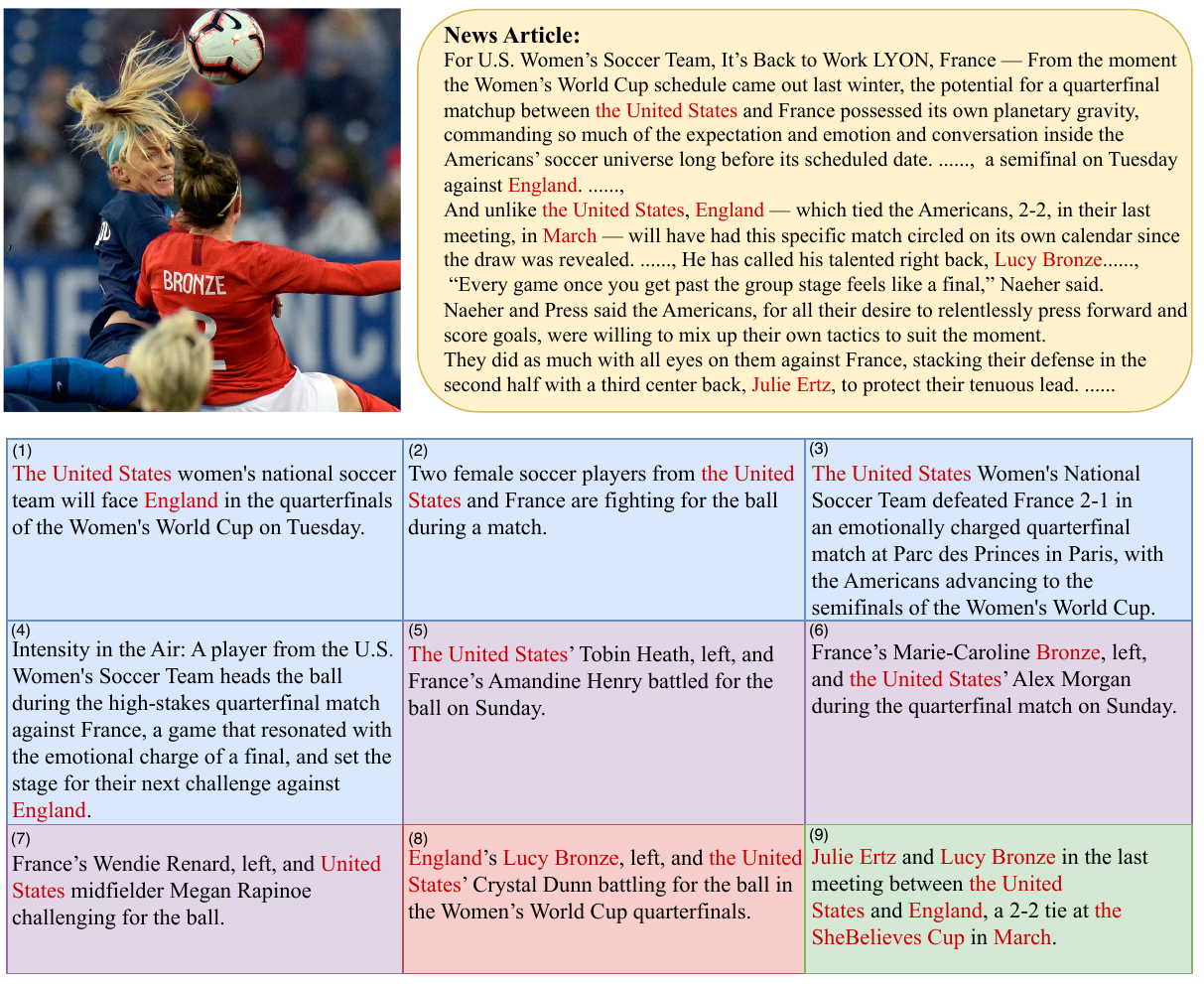}
    \caption{An example of news image caption generation. We present the news image, news article and the news image caption of one sample from NYTimes800k dataset. The news image captions are  respectively, generated in: (1) InstructBLIP in zero-shot setting. (2) LLaVA-v1.5 in zero-shot setting. (3) MiniGPT-v2 in zero-shot setting. (4) GPT-4V in zero-shot setting. (5) InstructBLIP (OSFT) on NYTimes800k. (6) LLaVA-v1.5 (OSFT) on NYTimes800k. (7) MiniGPT-v2 (OSFT) on NYTimes800k. (8) Our method, EAMA, on NYTimes800k. (9) is the ground-truth caption. We highlight Entities occurred in reference in \red{red}.}
    \label{fig:full_case}
\end{figure*}

We randomly select a sample from NYTimes800k dataset and generate captions using different models, as illustrated in Figure \ref{fig:full_case}. From the generated captions, it's evident that MLLMs under zero-shot settings can only produce common entity information such as "The United States" or "England", even when provided with the news image and associated text. Even the most powerful MLLM, GPT-4V \citep{openai2023gpt4} fails to generate specific entities related to the news image and mentioned in the news article, such as "Julie Ertz" and "Lucy Bronze."
After OSFT on NYTimes800k dataset, InstructBLIP still cannot generate the names of people in the news image. LLaVA-v1.5 successfully generates "Bronze" but overlooks "England." In contrast, our method generates both common entities and the names of people mentioned in the news article, further affirming the effectiveness of our approach in entity recognition. 

\begin{figure*}[tbp]
    \centering
    \includegraphics[width=0.97\textwidth]{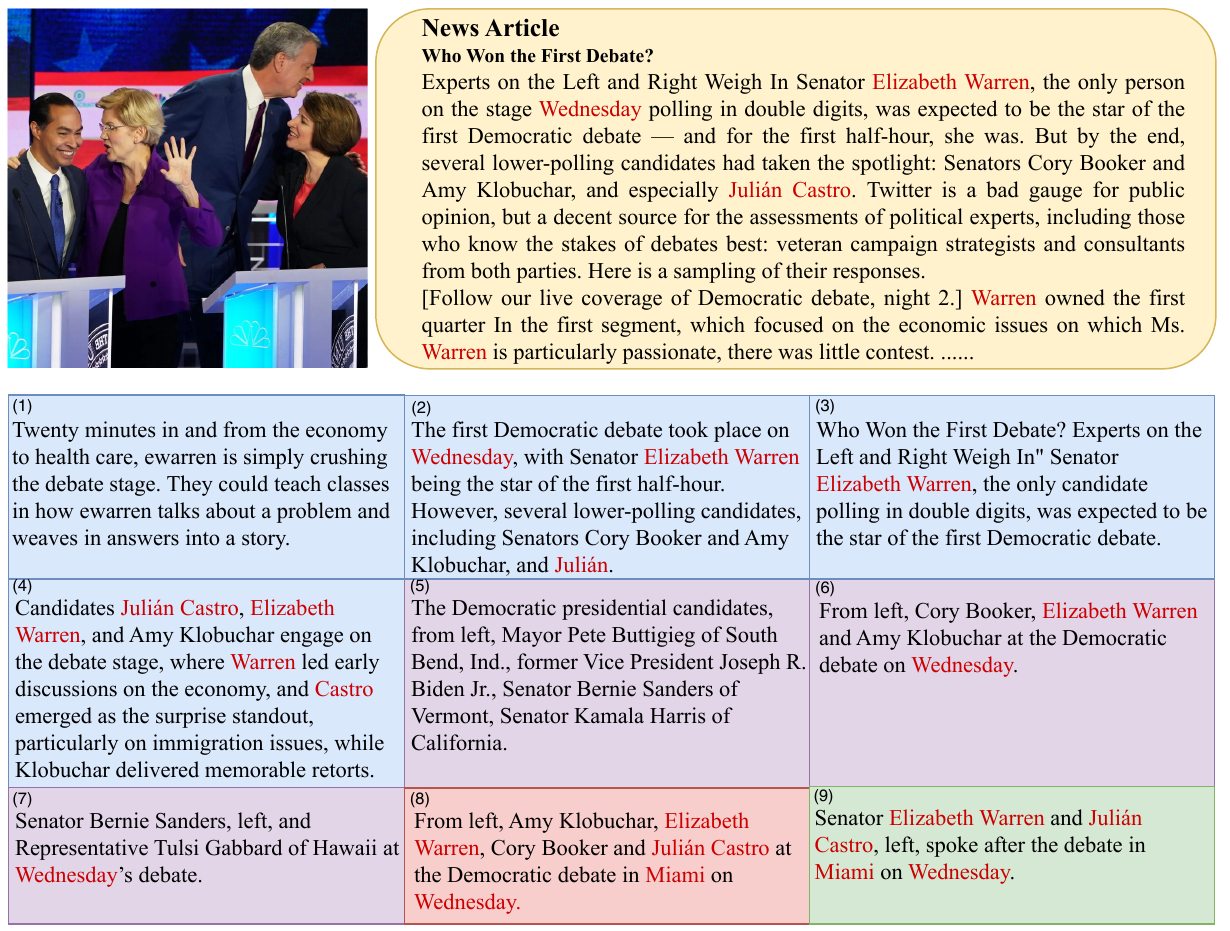}
    \caption{An example of news image caption generation. We present the news image, news article and the news image caption of one sample from NYTimes800k dataset. The news image captions are  respectively, generated in: (1) InstructBLIP in zero-shot setting. (2) LLaVA-v1.5 in zero-shot setting. (3) MiniGPT-v2 in zero-shot setting. (4) GPT-4V in zero-shot setting. (5) InstructBLIP (OSFT) on NYTimes800k. (6) LLaVA-v1.5 (OSFT) on NYTimes800k. (7) MiniGPT-v2 (OSFT) on NYTimes800k. (8) Our method, EAMA, on NYTimes800k. (9) is the ground-truth caption. We highlight Entities occurred in reference in \red{red}.}
    \label{fig:full_case2}
\end{figure*}

We also show another example to further illustrate the effectiveness of our method in figure \ref{fig:full_case2}. Our method achieves the best generation on entities and captions on both examples.

\end{document}